\documentclass{article}
\usepackage{PRIMEarxiv}
\usepackage{algpseudocode}
\usepackage{amsmath}
\usepackage{amsfonts}
\usepackage{amssymb}
\usepackage{multirow}
\usepackage{multicol}
\usepackage{graphicx}
\usepackage{caption}
\usepackage{subcaption}
\usepackage[T1]{fontenc}
\usepackage{comment}
\usepackage{booktabs}
\usepackage{pifont}
\usepackage[linesnumbered,ruled,vlined]{algorithm2e}
\usepackage{wrapfig}
\usepackage{xcolor}
\usepackage{graphicx}
\usepackage{xspace}
\usepackage{eso-pic}
\usepackage[T1]{fontenc}
\usepackage{mathptmx}
\usepackage{helvet,courier}
\usepackage{listings}
\newcommand{\xmark}{\ding{55}}%
\newcommand{\cmark}{\ding{51}}%

\title{S-JEA: Stacked Joint Embedding Architectures for Self-Supervised Visual Representation Learning}
\author{
	Alžběta Manová, Aiden Durrant and Georgios Leontidis \\
	Department of Computing Science \\
	University of Aberdeen \\
	Aberdeen, United Kingdom\\
	\texttt{\{a.manova.19, a.durrant.20, georgios.leontidis\}@abdn.ac.uk} \\
	}
\begin{document}

\maketitle

\begin{abstract}
The recent emergence of Self-Supervised Learning (SSL) as a fundamental paradigm for learning image representations has, and continues to, demonstrate high empirical success in a variety of tasks. However, most SSL approaches fail to learn embeddings that capture hierarchical semantic concepts that are separable and interpretable. In this work, we aim to learn highly separable semantic hierarchical representations by stacking Joint Embedding Architectures (JEA) where higher-level JEAs are input with representations of lower-level JEA. This results in a representation space that exhibits distinct sub-categories of semantic concepts (e.g., model and colour of vehicles) in higher-level JEAs. We empirically show that representations from stacked JEA perform on a similar level as traditional JEA with comparative parameter counts and visualise the representation spaces to validate the semantic hierarchies. 
\end{abstract}

\section{Introduction}
\label{sec:intro}
Over the past decade, deep learning has seen unprecedented growth and has become the main building block of AI systems. This has naturally emerged from the availability of large amounts of annotated data and benchmark datasets, such as ImageNet \cite{krizhevsky2017imagenet}. However, it has been well established that collecting and curating large annotated datasets is a laborious, time-consuming, and non-practical process in most cases. Still, most deep learning research has been primarily based on supervised methods, despite many real-life applications involving large amounts of unlabelled datasets. 

In the past couple of years, self-supervised learning (SSL) approaches have emerged as a way to make use of unlabelled data to pre-train large backbone CNN and Transformer models, and then use fewer labelled data to fine-tune the models for specific downstream tasks. Nowadays, SSL approaches perform competitively compared to their supervised counterparts across several downstream tasks \cite{gidaris2021obow,schiappa2022self}. This has been boosted by different architectures and a very comprehensive body of research \cite{bardes2021vicreg,doersch2015unsupervised,durrant2022hyperspherically,grill2020bootstrap,tian2020contrastive,tian2020makes,zbontar2021barlow,de2024object} that has proposed ways to pre-train deep learning models, such as ResNet \cite{he2016deep}, Capsule Neural Networks \cite{everett2024masked}, and Vision Transformers \cite{dosovitskiy2020image}, employing different types of SSL methods, primarily contrastive \cite{misra2020self,chen2020simple}, clustering \cite{caron2020unsupervised, yan2020clusterfit}, distillation \cite{gidaris2021obow, chen2021exploring}, information maximisation \cite{ermolov2021whitening, zbontar2021barlow}, and masking \cite{he2022masked,assran2023self}.

Recently, Image-based Joint-Embedding Predictive Architecture (I-JEPA) was proposed \cite{assran2023self}, a non-generative approach for self-supervised learning that learns to predict the embeddings of a signal \textit{y} from a compatible signal \textit{x}. The main block that enables this is a predictor network that is conditioned upon an additional variable \textit{z} to accommodate prediction.

\begin{figure}[t]
    \centering
    \includegraphics[width=.9\linewidth]{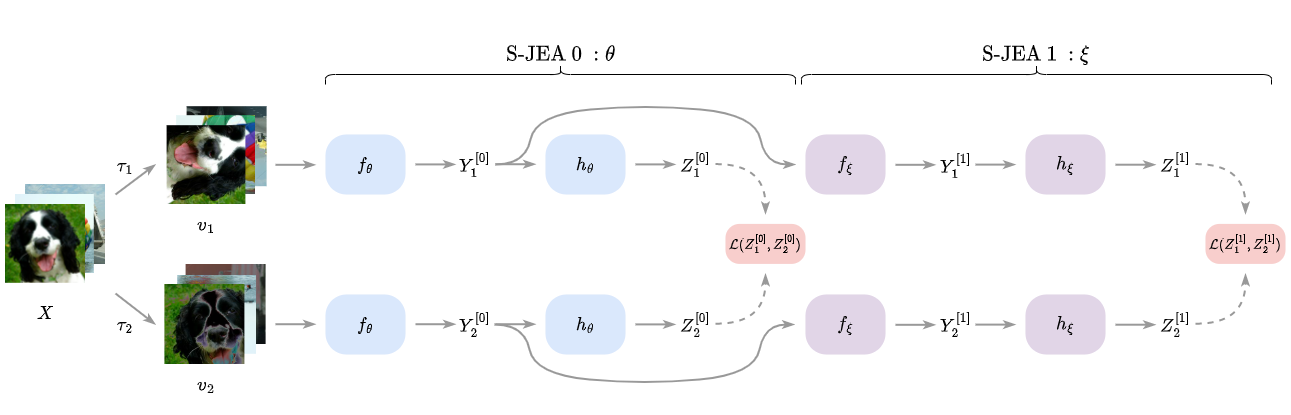}
    \vspace{0.5em}
    \caption{\textbf{S-JEA: Stacked Joint Embedding Architecture with 2 stacks.} A batch of images $X$ is transformed under two sets of transformations ($\tau_1$ and $\tau_2$) to produce two batches of views $V_1$ and $V_2$. The views are input to the first level encoder $f_\theta$ resulting in representations $Y^{[0]}_1$ and $Y^{[0]}_2$, for the first level stack the representations are expanded by $h_\theta$ producing embeddings $Z^{[0]}_1$ and $Z^{[0]}_2$. The second level stack takes the first level representations $Y^{[0]}_1$ and $Y^{[0]}_2$ and encodes them by $f_\xi$ to produce representations $Y^{[1]}_1$ and $Y^{[1]}_2$ which are expanded to form embeddings $Z^{[1]}_1$ and $Z^{[1]}_2$. For both stacked levels we use the VICReg loss applied at the embedding level of each stack, and the final loss is a weighted summation.}
    \label{SJEA}
\end{figure}

In this work, we propose stacked joint embedding architectures (S-JEA) for image tasks (Figure \ref{SJEA}) which learn hierarchical visual semantics by stacking joint embedding architectures. Learning hierarchies of semantics is a well-understood concept of representation quality \cite{bengio2013representation} and where learnt in separable and organised structures has shown to lead to improved downstream task performance, notably in few-shot learning \cite{zhang2022tree}. The learning of hierarchical semantics in vision tasks typically follows complex and computationally expensive methods such as hierarchical clustering \cite{yuan2023k, lin2022contrastive} or hyperbolic learning \cite{khrulkov2020hyperbolic}. Here we investigate if hierarchical semantics can be better learnt by performing self-supervision on the output representations of a lower-level self-supervised model by stacking encoders with the goal to learn abstract concepts at higher-level representations. We utilize the VICReg joint embedding architecture as the self-supervised method for all stacks given its good performance, simplicity, and understood mathematical underpinnings. 

To demonstrate the capabilities of S-JEA we empirically study the downstream task performance on the learnt representations at different stacked levels, visualise the representation space to interpret learnt hierarchies, perform a number of ablations, and analyse variations of loss propagation on the stack. The stacked joint embedding architecture presented in this work is largely general purpose, and can in theory, be applied to a number of joint embedding architectures such as I-JEPA\cite{assran2023self}, BYOL \cite{grill2020bootstrap} and SimCLR \cite{chen2020simple}.

\subsection{Motivation and Contributions}
As alluded to, we investigate whether learning representations from representations can lead to more abstract and thus, hierarchically structured visual semantic concepts. The intention is to first and foremost learn high-quality representations that are performant in downstream tasks, second is to investigate if learning higher-level representations from lower-level representations produces more abstract and hierarchically structured embedding spaces. The latter is motivated by \cite{lecun2022path} where we aim to investigate and move towards the ability of SSL methods to learn abstract semantic concepts by stacking whilst omitting the task of time-series state prediction. We analyse the above through an extensive empirical evaluation and various ablation studies, reporting on a variety of benchmark tasks and datasets. To summarise:

\begin{itemize}
    \item We present a method for stacked joint embedding architectures for self-supervised visual representation learning.
    \item We empirically demonstrate that our proposed method learns separable hierarchical semantic representations when stacking.
    \item S-JEA outperforms traditional architectures with comparative parameter counts on linear evaluation tasks.
\end{itemize}

\section{Related Work}
\subsection{Self-Supervised Learning}
Self-supervised visual representation learning aims to learn representations from a large set of unlabelled images such that the representations can be employed for a downstream task where labels are available. SSL learns without labels by the construction of proxy tasks that are informed by the data itself such as predicting an occluded region of the image \cite{pathak2016context, he2022masked}. The success of self-supervised learning can be attributed to the methodology of joint embedding architectures that task a Siamese network to produce representations that are invariant to transformations applied to the input image, this is typically done by maximising the similarity between the two representations \cite{chen2020simple, he2020momentum, bardes2021vicreg, zbontar2021barlow, grill2020bootstrap}. Such a formulation can lead to trivial solutions known as collapse, where output representations are identical regardless of the input. Contrastive methods \cite{chen2020simple} are one approach avoiding collapse by sampling negative views representations (representations that do not originate from the same image) and enforcing these to be dissimilar.  Although performant, contrastive methods require significant resources from large batch sizes \cite{chen2020simple} or memory banks \cite{he2020momentum} to ensure a sizeable sample of negative views. Instead, a number of alternatives to avoid collapse such as Barlow Twins \cite{zbontar2021barlow} and VICReg \cite{bardes2021vicreg} propose the use of covariance regularization on the representations to avoid collapsing solutions. This work primarily employs the latter VICReg as the base architecture, due to its simplicity both mathematically and computationally.

\subsection{Learning Hierarchies}
The purpose of stacking joint embedding architectures is fundamentally to explore the potential to learn abstract notions of representations such that a hierarchy of visual semantic concepts is formed. Learning hierarchies of visual semantics is not a new notion given its importance in tasks such as action recognition \cite{sun2009hierarchical, castro2022unsupervised} and large-scale image processing \cite{chen2022scaling, yuan2023k}. In addition to task-specific settings, models learning hierarchies have been shown to improve representation quality showing structured embeddings improve separability \cite{garg2022hiermatch, khrulkov2020hyperbolic}. This work is inspired by the propositions of \cite{lecun2022path} where we aim to first determine if the simplistic approach to stacking joint embedding architectures can learn meaningful, and abstract semantic hierarchies.

\section{Stacked Joint Embedding Architecture}
\subsection{Base JEA: VICReg}
Before proposing the stacking of Joint Embedding Architectures (JEA), we first define the model and pipeline for a single JEA stack. We employ the standardised form of VICReg JEA \cite{bardes2021vicreg} as our base given its computation and mathematical simplicity which provides strong results that are easily reproduced. 

Given a batch of images $X$ sampled from dataset $\mathcal{D}$ and two transformations $\tau_1, \tau_2$ sampled from $\mathrm{T}$, two corresponding views of each image $i$ in the batch are produced $v_1=t(i)$ and $v_2=t'(i)$ resulting in batches $V_1$ and $V_2$. The views are encoded into their representations $Y_1=f_{\theta}(V_1), Y_2 = f_{\theta}(X_2)$ before being mapped to their embeddings $Z_1=h_{\theta}(Y_1), Z_2=h_{\theta}(Y_2)$.

The identically parameterised Siamese encoders $f_{\theta}$ and their projectors $h_{\theta}$ are trained to minimise the weighted summation of the invariance, variance, and covariance given by:

\begin{equation}
    \ell(Z_1, Z_2) = \lambda s(Z_1, Z_2)+ \mu [v(Z_1) + v(Z_2)] + \upsilon [c(Z_1) + c(Z_2)] ,
    \label{eq:vicreg}
\end{equation}

where $s$,$v$, and $c$ are the invariance loss, variance, and covariance terms respectively, while $\lambda$, $\mu$, and $\upsilon$ are the associated weighting of each term. For more details, we refer the reader to the original work \cite{bardes2021vicreg}.

\subsection{Stacking JEAs}
The motivation for stacking JEAs is to learn higher-level abstract representations of lower-level representations. To achieve this we propose to apply a second VICReg JEA ($f_\xi$) to the output representations, $Y^{[0]}_1$ and $Y^{[0]}_2$, of the first JEA ($f_\theta$), this is visually depicted in Figure \ref{SJEA} and algorithmically defined in Algorithm.\ref{alg:sjea}.

The encoders $f_\theta$ and $f_\xi$ of each stack are both ResNet-18 \cite{he2016deep} convolutional neural networks, with the subsequent stack modified to accept the size and channels of the representations of the previous encoder output. Typically, the output of the ResNet is average-pooled to produce a representation vector, however, we find in practice that passing the representation before the pooling layer to the following stack improves performance as no resizing and interpolation is required. 
We maintain this pooling layer when applying the projector networks $h_\xi$ for each stack. The projector networks $h_\theta$ and $h_\xi$ are comprised of three fully-connected layers the first two followed by batch normalization (BN) \cite{ioffe2015batch} and ReLU non-linear activation.  

Training S-JEA is done in a stack-wise manner where the VICReg objective (Equation.\ref{eq:vicreg}) is minimised for the encodings for each stack and the loss propagated back through all previous stacks. The final objective is the sum of the VICReg losses for each stack encoding. We find empirically that the optimal weighting coefficients ($\lambda = 25$, $\mu = 25$, $\upsilon = 1$) of the loss terms are those proposed in the original work \cite{bardes2021vicreg} for all stacks.

The aforementioned formulation of S-JEA is empirically found to be the optimal configuration, we further demonstrate the impact of the projector head in \ref{sec:proj}, explore encoder depth in contrast to stacking \ref{sec:depth}, and visualise the semantic hierarchies of stacked encoder representation spaces in \ref{sec:viz}.

\begin{algorithm}
\caption{PyTorch styled pseudo code for S-JEA}\label{alg:sjea}
\begin{lstlisting}
# f1, f2: encoder networks; 
# h1, h2: expander networks;
# lambda, mu, nu: loss term coefficients; 
for x in dataloader:
    # Augment
    x_1, x_2 = transform(x)
    
    # Stack 0
    y0_1, y0_2 = f1(x_1), f1(x_2) # representations
    z0_1, z0_2 = h1(y0_1), h1(y0_2) # embeddings
    # calculate loss
    loss0 = lambda * i_loss(z0_1, z0_2) + mu * v_loss(z0_1, z0_2) + nu * c_loss(z0_1, z0_2)

    # Stack 1
    y1_1, y1_2 = f2(y1_1), f2(y1_2) # representations
    z1_1, z1_2 = h2(y1_1), h2(y1_2) # embeddings
    # calculate loss
    loss1 = lambda * i_loss(z1_1, z1_2) + mu * v_loss(z1_1, z1_2) + nu * c_loss(z1_1, z1_2)
    
    loss = loss0 + loss1
\end{lstlisting}
\end{algorithm}
\vspace{-2em}
\section{Empirical Results}
In this section, we empirically report the performance of the learnt representations at each stack's encoder. Following this we conduct a series of ablations studying the impact on S-JEA configurations of the encoder \ref{sec:depth}, specifically focusing on the role of the projection head in the stack\ref{sec:proj}. Finally, we visualize the representation space for the stacked encoder showing its unique clustering of sub-semantic concepts \ref{sec:viz}.
\subsection{Linear Evaluation}
We evaluate using linear evaluation the quality of representations learnt during self-supervised pre-training of each ResNet-18 encoder on CIFAR-10 and STL-10 datasets for 200 epochs and 300 epochs respectively. During this procedure we train a linear classifier on top of the frozen, pre-trained encoder networks after each stack individually for 200 epochs (300 epochs for STL-10) this is aligned with the linear evaluation protocol described in \cite{bardes2021vicreg, grill2020bootstrap}. Specifically, we discard the projector for the encoder we are evaluating, and when evaluating stacked encoders the subsequent encoders are also frozen.

\begin{table}[h!]
    \centering
    \begin{tabular}{l |c |c |c |c |c}
        \toprule
         \multirow{ 2}{*}{Method} & \multirow{ 2}{*}{Encoder Stack Level} & \multicolumn{2}{c}{CIFAR-10} & \multicolumn{2}{|c}{STL-10}\\
          &  & top-1 \% & $k$-NN \% & top-1 \% & $k$-NN \%\\
         \midrule
         VICReg & No Stack & 80.5 & 83.2 & 75.9 & 75.1 \\  
         \midrule
         S-JEA & 0 & 80.8 & 83.4  & 76.5 & 75.3 \\
         S-JEA & 1  & 81.6 & 81.1 & 72.3 & 71.6 \\   
         \midrule
    \end{tabular}
    \vspace{1em}
    \caption{\textbf{CIFAR-10 and STL-10 Linear Evaluation:} Baseline and stacked encoder trained for 200 epochs for CIFAR-10 and 300 for STL-10, a linear classifier trained on the frozen representations from each encoder. We report top-1 and $k$-NN accuracy (\%).}
    \label{tab:linear}
\end{table}

\begin{figure}[t]
    \centering
        \begin{subfigure}{0.33\textwidth}
            \centering
            \includegraphics[width=0.9\textwidth]{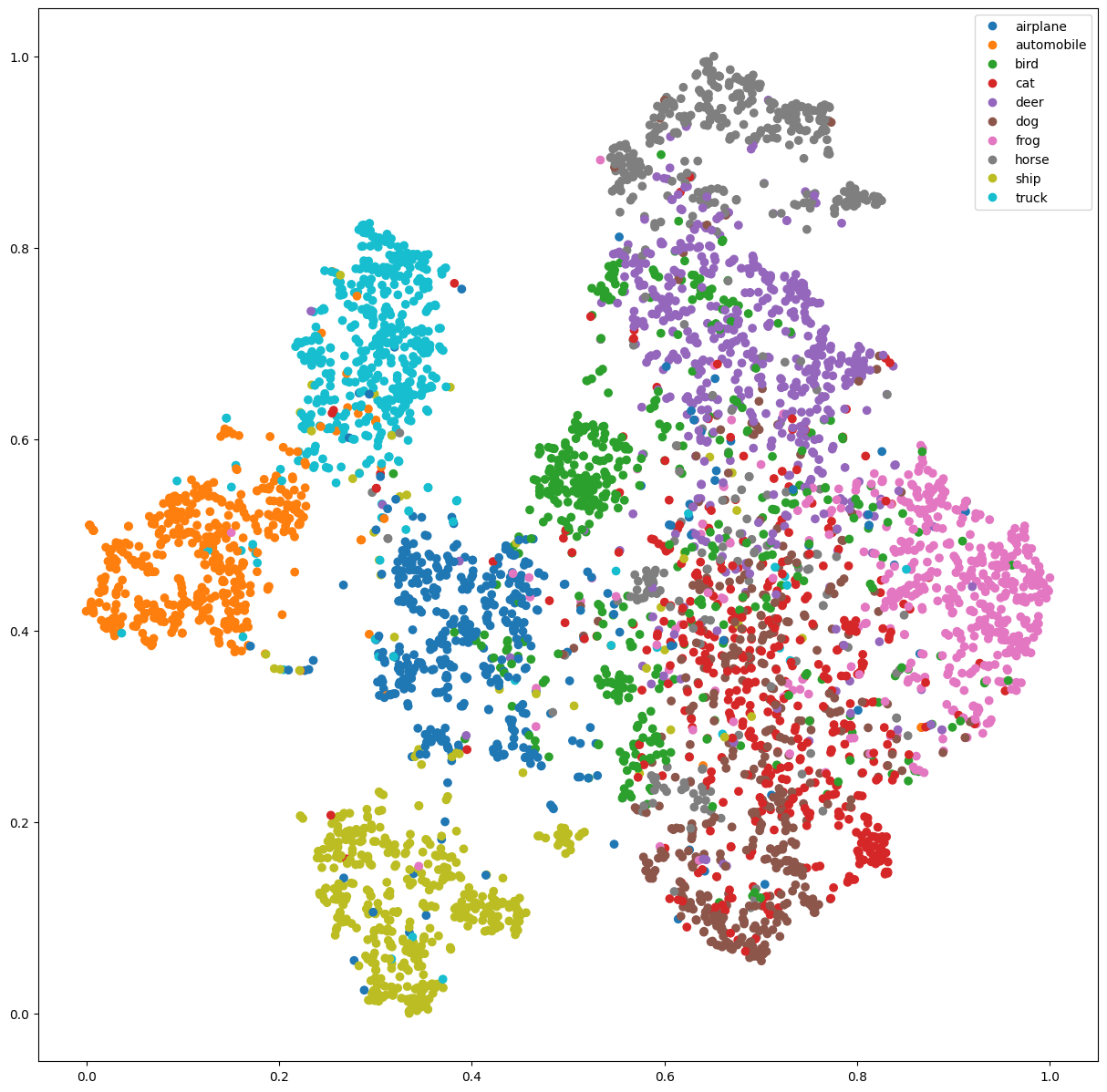}
            \caption{VICReg Embeddings}
            \label{fig:vicpretrain}
            \centering
        \end{subfigure}%
        \begin{subfigure}{0.33\textwidth}
            \centering
            \includegraphics[width=0.9\textwidth]{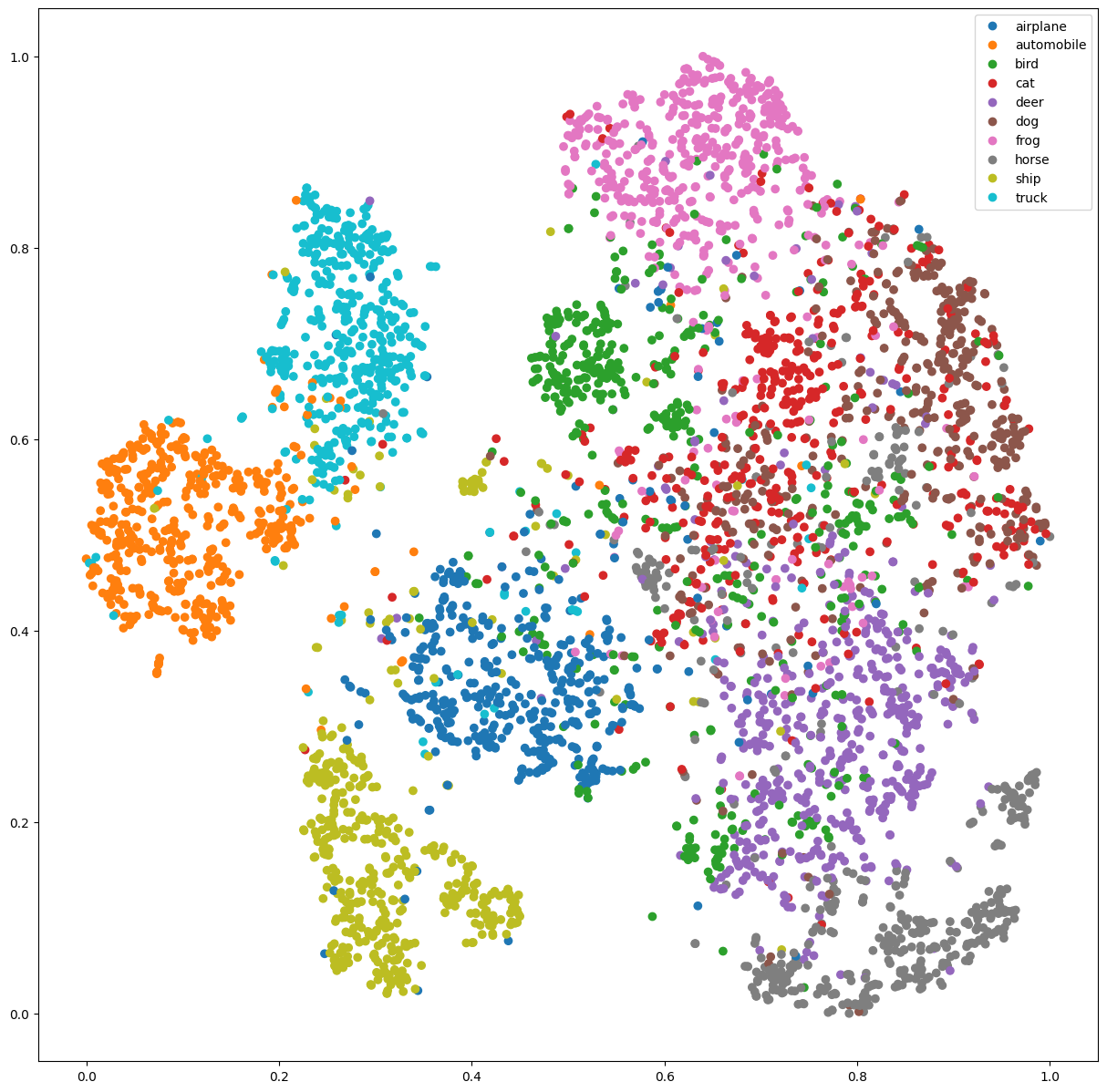}
            \caption{S-JEA: Stack 0 Embeddings}
            \label{fig:svic0pretrain}
            \centering
        \end{subfigure}%
        \hfill
        \begin{subfigure}{0.33\textwidth}
            \centering
            \includegraphics[width=0.9\textwidth]{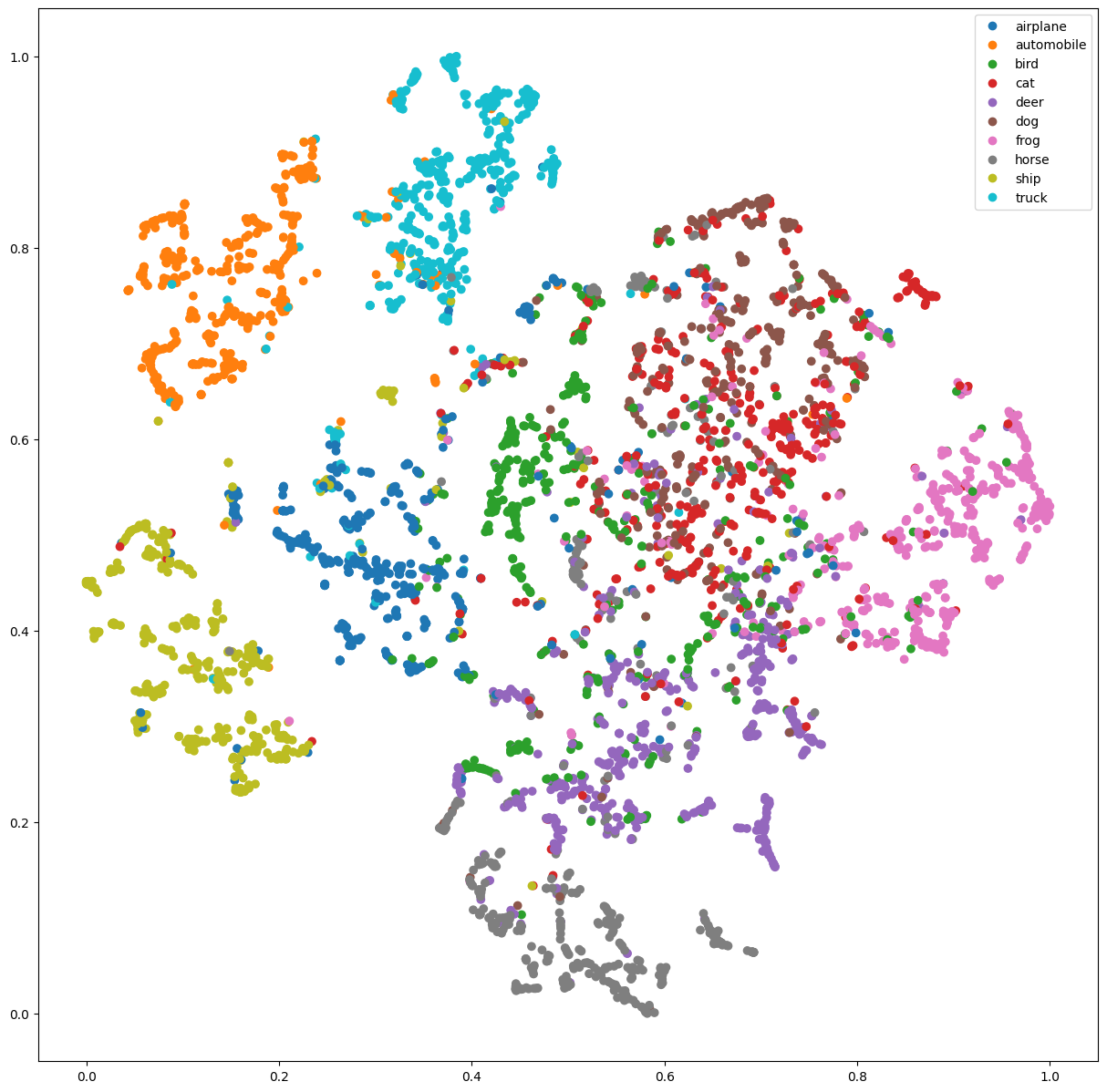}
            \caption{S-JEA: Stack 1 Embeddings}
            \label{fig:svic1pretrain}
            \centering
        \end{subfigure}
    \vspace{1em}
    \caption{\textbf{Visualisation of Learned Representations on CIFAR-10.} t-SNE plots of the CIFAR-10 test set representations frozen encoders trained by VICReg and S-JEA at each stack.}
    \centering
    \label{fig:tsne_cifar}
\end{figure}

Table \ref{tab:linear} reports the top-1 accuracy (\%) and K-Nearest Neighbour ($k$-NN) accuracy (\%) by comparing our method and the baseline non-stacked approach. Our results are reported as the average over 3 randomly initialised runs on the frozen encoder networks. We observe from the results that S-JEA outperforms base VICReg on both datasets, for linear evaluation and $k$-NN. For CIFAR-10 it is demonstrated that stacking the encoders provides a reasonable performance improvement over baseline VICReg, with stack 0 performing comparatively as expected. We can visually observe from t-SNE plots in Figure \ref{fig:tsne_cifar} that the first encoders CIFAR-10 representations (Figure \ref{fig:svic0pretrain}) are largely similar to the baseline VICReg model (Figure \ref{fig:vicpretrain}), whilst the representations of the subsequent stacked encoder exhibit strong sub-clusters within the semantic clusters (Figure \ref{fig:svic1pretrain}). This phenomena is discussed in more detail in \ref{sec:viz}. Interestingly for STL-10 we observe that stack 0 encoder outperforms the subsequent stack, we conjecture as to why we observe this peculiarity in the following sections. 

\subsection{Comparing Deep Vs. Stacked}\label{sec:depth}
The addition of the second encoder stack doubles the number of trainable parameters, therefore posing the question if the improved representations quality is simply a product of a more expressive network and not the stacking dynamics. Table \ref{tab:parameters} shows the performance of the baseline VICReg models with different encoder parameter counts, notably, we employ a deeper ResNet-50 containing approximately 23 million trainable parameters which are comparable to the parameter count of our proposed stacked joint embedding architecture. 

\begin{table}[h!]
    \centering
        \begin{tabular}{ l|c|c|c|c } 
            \toprule
        Method & Model & Parameters (M) & top-1 \% & $k$-NN \%\\
            \midrule
        VICReg & ResNet-18 & 11.6 & 80.5 & 83.2 \\
        VICReg & ResNet-50 & 23.0 & 82.0 & 83.4\\
            \hline
        S-JEA : 1 & ResNet-18 & 23.2 & 80.8 & 83.4 \\
        S-JEA : 2 & ResNet-18 & 23.2 & 81.6 & 81.1 \\
        \bottomrule
        
    \end{tabular}
    \vspace{1em}
    \caption{\textbf{Parameter Counts:} Baseline and stacked encoder trained for 200 epochs on CIFAR-10 for varying model sizes, a linear classifier trained on the frozen representations from each encoder. We report top-1 accuracy (\%). ResNet-50 offers similar parameter counts to a 2-stack S-JEA for the evaluation of stacking dynamics as opposed to parameter count.}
    \centering
    \label{tab:parameters}
\end{table}

The top-1 accuracy indicates that the proposed S-JEA performs comparably to the baseline VICReg with model ResNet-50, as reported in Table \ref{tab:parameters}. Even though we can observe a minor decrease in top-1 accuracy for S-JEA, the recorded values for $k$-NN  confirm that stacking does not reduce the quality of the architecture's performance considering stack 0. It can be assumed that the accuracy of stack 0, is indeed improved by the addition of trainable parameters. On the other hand, stack 1 still exhibits acceptable performance. We speculate that while general performance is improved as a result of the increase in the parameter count, the stacking approach also delivers hierarchical classification and learns representations that are omitted from the usual JEA as shown in \ref{sec:viz}.

\subsection{Projection Head} \label{sec:proj}
The projection head is a vital component in the learning of good representations for downstream tasks \cite{chen2020simple}. Applied on the representations of the encoder networks to produce embeddings with which the loss is performed, the projection head has been shown to enable significant performance gains \cite{bordes2022guillotine} when there exists misalignment between training and downstream tasks, naturally present in SSL. Importantly, the retention of the projector in downstream tasks is limiting in performance given the reduced generalisation capabilities of the embeddings learnt under the SSL objective. Therefore, in the case of stacking JEAs a stacked encoder can receive the embeddings of the projector as input and we then investigate if the stacked encoder that receives the projector embeddings as input is less performant on the downstream tasks.

\begin{figure}
	\begin{center}
		\includegraphics[scale=0.45]{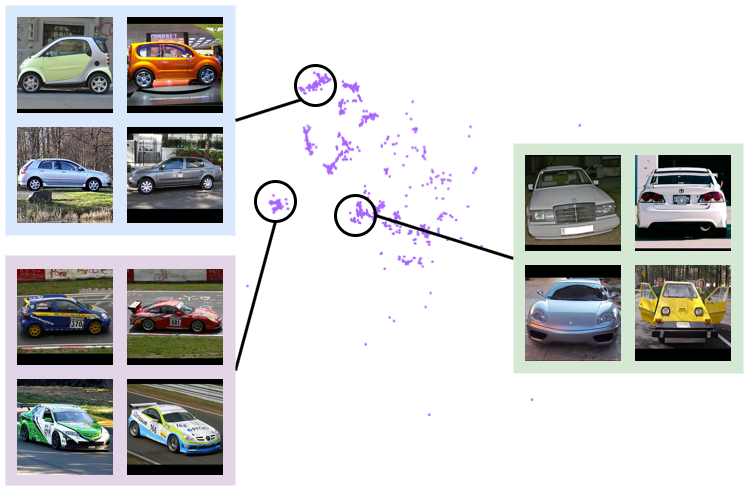}
		\vspace{1em}
		\caption{\textbf{Visualisation of Semantic Sub-Clusters of S-JEA.} t-SNE plot of STL-10 test set representations of the frozen pre-trained stacked encoder. The representations shown correspond to the semantic class label `\textit{cars}'. Of the three identified sub-clusters, the \textit{top} shows image representations pertaining to side view poses of cars, the \textit{bottom right} consists of forward facing pose, whilst the \textit{bottom-left} sub-cluster contains all race cars.}
		\label{fig:hierarchies}
	\end{center}
\end{figure}

\subsection{Semantic Hierarchies} \label{sec:viz}
For a joint embedding architecture with two levels of encoding, there are four configurations to be investigated for the projection head. Table \ref{tab:projector} reports linear evaluation performance for each encoder with and without the projector.

\begin{table}[h!]
    \centering
    \begin{tabular}{l |c |c |c |c}
        \toprule
          & \multicolumn{4}{c}{Projector Configuration}\\
         \midrule
         Stack 0 Projector & \xmark & \cmark& \xmark & \cmark \\
         Stack 1 Projector & \xmark & \xmark & \cmark & \cmark \\
         \midrule
         \multicolumn{5}{c}{Stack 0 Encoder}\\
         \midrule
         Accuracy \% & 83.3 & 80.8 & 83.2  & 80.09  \\
         $k$-NN \%  & 81.6 & 83.4 & 81.2  & 82.5\\
         \midrule
         \multicolumn{5}{c}{Stack 1 Encoder}\\
         \midrule
         Accuracy \% & 80.5 & 81.6 & 80.8 & 79.9 \\
         $k$-NN \%   & 80.1 & 81.1 & 80.3 & 80.2 \\
    \end{tabular}
    \vspace{1em}
    \caption{\textbf{Projector Configuration:} linear evaluation of the frozen encoders when pre-trained with and without the projector head at each stack.}
    \label{tab:projector}
\end{table}

The results in Table \ref{tab:projector} show that the projector network is not as fundamental of a component in S-JEA compared to the baseline VICReg model \cite{bordes2022guillotine}, where the performance does not degrade significantly in the absence of such projectors. However, we conjecture that in the cases where no projector is present on stack 0, the second stack effectively behaves as the projector network and as a result when dropped for linear evaluation maintains the performance of baseline methods. This aligns with the findings of \cite{garrido2022duality} in which downstream performance increases with the dimensionality of the projector. This is an important distinction to note when training S-JEA to ensure that stacked joint embedding architectures can be effective representation learners. One possible solution that we pose for future work is to ensure that each stack remains unique by introducing a stop gradient, akin to \cite{grill2020bootstrap}, between the stacks, this ensures that the loss propagation is stack-wise.

\begin{figure}
\vspace{-2.5em}
  \begin{center}
    \includegraphics[width=0.85\textwidth]{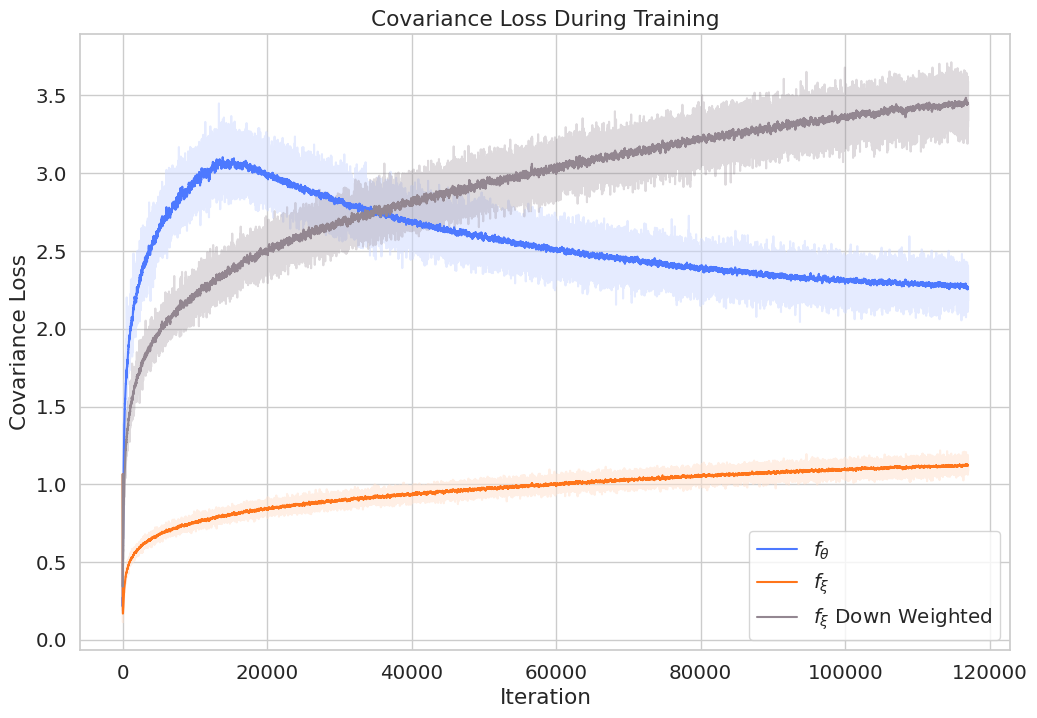}
    \vspace{1em}
    \caption{\textbf{Covariance Loss Term During Training.} Covariance loss term for both first level (\textit{blue}) and higher level stacked encoders (\textit{orange}) during training. The covariance loss term when weighted down is also presented (\textit{grey}). }
        \label{fig:cov_loss}
  \end{center}
\end{figure}

The above results give a good indication regarding the overall ability of S-JEA to learn good image representations. However, the intention of S-JEA is not to only improve downstream task performance but to investigate the ability to learn separable hierarchical semantic concepts in higher-level stacks. The t-SNE plots in Figure \ref{fig:tsne_cifar} show how semantic sub-clusters of representations form from the stacked encoders in comparison to the baseline VICReg. Figure \ref{fig:hierarchies} visualises these sub-clusters, and we show that distinct sub-clusters for the semantic class `\textit{car}' corresponding to pose (e.g. side facing, and forward facing) and visual appearance (e.g. race cars) are formed. It is worth noting that the t-SNE diagrams in Figure \ref{fig:svic1pretrain} and Figure \ref{fig:hierarchies} exhibit typical signs of feature collapse \cite{gupta2022understanding}. Yet, when observing the covariance term of the VICReg loss during training (Figure \ref{fig:cov_loss}) we see that, in fact, the covariance of embedding dimensions remains low and far lower than with covariance term down-weighted, thus indicating that information collapse is not the driver of this visual observation.

The sub-clusters visualised are present for all semantic classes, however, in comparison to baseline and low-level encoder representations we observe significantly more overlap of the labelled semantic classes in higher abstraction concept sub-clusters (i.e. pose, shape, colour, etc.). This phenomenon, we conjecture, is one factor describing the disparity in downstream linear evaluation performance for higher-level encoder representations but higher $k$-NN accuracy where the clusters are not restricted by the task-defined labels. Further investigations could lead to promising research directions investigating such findings and evaluating performance on a variety of downstream tasks.

\section{Conclusion and Future Work}
Stacked Joint Embedding Architectures (S-JEA) presented in this work provide key insights towards learning abstract hierarchical semantic representations of visual features. We demonstrate that S-JEA is viable in learning good quality representations from images that outperform traditional methods in linear evaluation benchmarks even when parameter counts are comparative. Furthermore, we show that the embeddings learnt in stacked layers exhibit high numbers of sub-clusters that are semantically similar confirmed through manual investigation, leading to improved performance in downstream tasks.
 We argue that future investigations of S-JEA could lead to innovative research into the embedding structure and semantic hierarchies, and potentially lead to vastly improved representation quality and interpretability.

\bibliographystyle{unsrt}

\end{document}